\documentclass{egpubl}
 
\JournalPaper         
\usepackage[T1]{fontenc}
\usepackage{dfadobe}  
\usepackage[labelfont=bf]{caption}

\usepackage[normalem]{ulem}

\usepackage[dvipsnames]{xcolor} 
\newcommand{\hl}[1]{#1} 

\usepackage{booktabs,ragged2e}

\usepackage[flushleft]{threeparttable}

\usepackage{amssymb}

\usepackage{caption}
\usepackage{booktabs}
\usepackage{multirow}

\usepackage{amsmath}

\usepackage{cite}  
\BibtexOrBiblatex
\electronicVersion
\PrintedOrElectronic
\ifpdf \usepackage[pdftex]{graphicx} \pdfcompresslevel=9
\else \usepackage[dvips]{graphicx} \fi
\graphicspath{{figures/}{pictures/}{images/}{./}} 

\usepackage[absolute,overlay]{textpos}
\usepackage{pbox}
\usepackage{everypage}

\newcommand{\stamp}[1][© 2023 The Eurographics Association and John Wiley \& Sons Ltd. This is the author's version of the article that has been published in Computer Graphics Forum. The final version of this record is available at: \href{https://doi.org/10.1111/cgf.14726}{\color{blue}10.1111/cgf.14726}]{%
\begin{textblock*}{140mm}(37mm,270mm)
\centering%
\small%
\emph{#1}%
\end{textblock*}%
}

\AddEverypageHook{
  \stamp
}

\usepackage{cleveref}
\renewcommand{\autoref}{\Cref}

\usepackage{egweblnk}

\usepackage{longfbox}

\newfboxstyle{tight2}{padding=2pt,margin=0pt,baseline-skip=false}%

\definecolor{ColT1}{HTML}{1f77b4}
\definecolor{ColT2}{HTML}{ff7f0e}
\definecolor{ColT3}{HTML}{2ca02c}
\definecolor{ColT4}{HTML}{d62728}
\definecolor{ColT5}{HTML}{9467bd}
\definecolor{ColT6}{HTML}{8c564b}
\definecolor{ColT7}{HTML}{e377c2}
\definecolor{ColT8}{HTML}{7f7f7f}
\definecolor{ColT9}{HTML}{bcbd22}
\definecolor{ColT10}{HTML}{17becf}

\newlength{\boxh}
\title[HardVis: Visual Analytics to Handle Instance Hardness Using Undersampling and Oversampling Techniques]%
      {HardVis: Visual Analytics to Handle Instance Hardness Using Undersampling and Oversampling Techniques}

\author[Chatzimparmpas et al.]
{\parbox{\textwidth}{\centering A. Chatzimparmpas$^{1}$\orcid{0000-0002-9079-2376}, F. V. Paulovich$^{2}$\orcid{0000-0002-2316-760X}, and A. Kerren$^{1,3}$\orcid{0000-0002-0519-2537}}
        \\
{\parbox{\textwidth}{\centering $^1$Department of Computer Science and Media Technology, Linnaeus University, Sweden\\
$^2$Department of Mathematics and Computer Science, Eindhoven University of Technology, The Netherlands\\
$^3$Department of Science and Technology, Link\"{o}ping University, Sweden}}
}

\begin{document}

\stamp

\maketitle

\begin{abstract}
   Despite the tremendous advances in machine learning (ML), training with imbalanced data still poses challenges in many real-world applications. Among a series of diverse techniques to solve this problem, sampling algorithms are regarded as an efficient solution. However, the problem is more fundamental, with many works emphasizing the importance of instance hardness. This issue refers to the significance of managing unsafe or potentially noisy instances that are more likely to be misclassified and serve as the root cause of poor classification performance.
%
This paper introduces HardVis, a visual analytics system designed to handle instance hardness mainly in imbalanced classification scenarios. Our proposed system assists users in visually comparing different distributions of data types, selecting types of instances based on local characteristics that will later be affected by the active sampling method, and validating which suggestions from undersampling or oversampling techniques are beneficial for the ML model. Additionally, rather than uniformly undersampling/oversampling a specific class, we allow users to find and sample easy and difficult to classify training instances from all classes. Users can explore subsets of data from different perspectives to decide all those parameters, while HardVis keeps track of their steps and evaluates the model's predictive performance in a test set separately. The end result is a well-balanced data set that boosts the predictive power of the ML model.
The efficacy and effectiveness of HardVis are demonstrated with a hypothetical usage scenario and a use case. Finally, we also look at how useful our system is based on feedback we received from ML experts.

   \makeatletter

	\def\customclassification{\vskip 5.5pt\par\reset@font\rmfamily}
	\def\endcustomclassification{\relax}
	\makeatother

	\begin{customclassification}
	\end{customclassification}
\end{abstract} 

\section{Introduction} \label{sec:intro}{%
  In machine learning (ML), \emph{easy to classify instances} are those for which ML models have a high probability of predicting the correct class label, whereas the opposite is true for the \emph{difficult to classify instances}~\cite{Yu2021BIDI}. The assessment of \textbf{instance hardness} can reveal useful information about the boundaries of ML capabilities~\cite{Prudencio2015Analysis}. \hl{Instance hardness is a common problem that even inspired the creation of well-known boosting algorithms~\cite{Yu2021An}, such as AdaBoost~\cite{Freund1999A}. It can also highlight when and where human intervention is required to resolve data-related issues.} The ultimate goal of such a procedure is to identify misclassified instances and interpret why this has happened~\cite{Castor2014Cost}, as well as improve predictive performance~\cite{Smith2014Instance}. \hl{This scenario is where visual analytics (VA) approaches are considered as a possible solid solution~\cite{Wu2022In} with many recent works focusing on problematic subsets of data for the interpretation and performance boost of ML models~\cite{Collaris2022StrategyAtlas,Zhang2022SliceTeller}.} However, the classification problem becomes significantly more complex when the data set contains both \emph{class overlap} and \emph{class imbalance}. There are many problems~\cite{Rao2006Data,Wei2013Effective,Herland2018Big,Cieslak2006Combatting,Kubat1998Machine} in which the minority class---composed of mostly unsafe instances such as borderline examples, rare cases, and outliers---is of great interest~\cite{Napierala2016Types}. \hl{A medical diagnosis task of detecting ill patients within a healthy majority is an example that illustrates the great importance of imbalanced data problems.} Learning from such unbalanced data sets can be difficult because most models will theoretically attain high accuracy by merely predicting the majority class~\cite{Stefanowski2016Dealing}.

There are two fundamental methodologies to deal with these kinds of imbalance problems: \emph{data-level} and \emph{algorithm-level approaches}~\cite{Krawczyk2016Learning}. The first method utilizes preprocessing strategies in order to balance the training set. The second method aims at determining what causes a certain ML model to fail in imbalanced circumstances and addressing those flaws to create new robust ML models~\cite{Cano2013Weighted,Czarnecki2017Extreme}. \emph{Ensemble approaches} have also grown in popularity, as they allow for a fusion of model combinations and the usage of one of the methodologies discussed above~\cite{Ksieniewicz2017Paired,Wozniak2014A}. In this paper, we solely focus on the data-level approaches because they are not entangled to a specific ML algorithm; and they remain as an underrepresented category without the support of VA solutions~\cite{Chatzimparmpas2020A,Chatzimparmpas2020The,Yuan2021Survey}. \hl{These approaches perform data sampling that refers to either undersampling or oversampling techniques. The former removes instances from the training set, while the latter generates synthetic/artificial instances from the already existing data to balance the class distribution.}

With regard to \emph{undersampling}, two advanced techniques for concurrently eliminating and maintaining instances are: one-sided selection (OSS)~\cite{Kubat97Addressing} and neighborhood cleaning rule (NCR)~\cite{Laurikkala2001Improving}. The goal here is to remove ambiguous points on the class boundary and, at the same time, keep any nonredundant examples far from the decision boundary. On the other hand, a frequently used \emph{oversampling} algorithm is the synthetic minority oversampling technique (SMOTE)~\cite{Chawla2002SMOTE} that comes with several drawbacks. One of those is the uniform approach to oversampling which considers all minority instances equally important. To deal with this flaw, adaptive synthetic (ADASYN)~\cite{He2008ADASYN} was invented that dynamically determines which cases may represent a greater challenge for an ML model, thus oversampling instances around class borders. A non-trivial issue with these algorithms is that they require the exploration of specific parameters from the user side. The common ground in all these techniques is the k-value that should be set for the k-nearest neighbors (KNN) algorithm~\cite{Altman1992Introduction,Fix1989Discriminatory}. Depending on this critical value, more or fewer instances will be removed or used for artificial addition, which might cause harm to the predictive performance of the ML model under training. \hl{For example, in an imbalanced healthcare data scenario, a data analyst who blindly trusts one of the previous heuristic-based approaches for undersampling and chooses a high k-value will eventually remove so many healthy patients (belonging to the majority class), leading to a balanced training set but with a significant loss of critical data for generalizing when the system will be put into production.} Tuning those parameters is not straightforward to be automated since there are multiple ways on how to combine undersampling and oversampling\hl{; thus, making room for human-centric solutions such as interactive visualizations that facilitate human exploration and domain knowledge injection into this complex problem.} Furthermore, the local characteristics of each instance are at least equally important as the global extracted patterns, which are usually investigated with automated methods~\cite{Ramentol2015IFROWANN}. Consequently, a remaining open question is: \textbf{(RQ1)} for a given data set, how can visualization assist users in deciding the optimal parameters for the undersampling and oversampling techniques?

Another challenge related to the previous one is to identify common \emph{local characteristics} of the instances in order to classify them into data types, as in the work of Napierala and Stefanowski~\cite{Napierala2016Types} that acknowledges four types of data: \emph{safe, borderline, rare,} and \emph{outliers} (SBRO in short). As described before, depending on the selected k-value, the distribution of instances in those types is subject to change~\cite{Skryjomski2017Influence}. Outliers can account for a sizable fraction of a class, especially in minority groups; as a result, in some data sets, they may even predominate~\cite{Napierala2016Types}. It is dangerous to treat outliers as noise and utilize noise-handling approaches such as relabeling or eliminating them from the learning set without extensively analyzing them~\cite{Xiang2019Interactive,Bauerle2020Classifier}. Separating noise from outliers is a necessary but non-trivial task~\cite{Salgado2016Noise}. \hl{If we consider the previously established example, a data analyst will receive various distributions of SBRO instances (i.e., separations of patients) depending on the k-value selected for splitting the data with KNN into these four data types, where some combinations will lead to more outliers that could be potentially treated as noisy data compared to others.} Moreover, rare cases exist in several data sets~\cite{Ravindran2011Learning}. This indicates that class difference is not the only source of difficulties when dealing with unbalanced data, but local characteristics of each class are also essential~\cite{Napierala2016Types}. This problem is partially addressed with upgraded versions of SMOTE and hybrid algorithms. For example, Borderline-SMOTE~\cite{Han2005Borderline} focuses on oversampling cases that are near to class boundaries. Safe-Level-SMOTE~\cite{Bunkhumpornpat2009Safe} allocates weights to instances based on how ``safe'' they are from the majority class influence, and it uses these weights to guide the introduction of artificial examples. Additionally, selective preprocessing of imbalanced data (SPIDER)~\cite{Napierala2010Learning} focuses on highlighting problematic cases, particularly those that overlap with the majority class. Nevertheless, it would be better to dynamically adjust this ratio based on the exploration of local data features and the varying density of examples. In such dynamic approaches, evaluating several types of data could be useful~\cite{Napierala2016Types}. Thus, a question that arises is: \textbf{(RQ2)} \hl{which algorithmic suggestions should users accept based on the visual analysis of particular SBRO areas or even whole regions?}

In this paper, we present a VA system, called \textsc{HardVis}, that incorporates undersampling and oversampling techniques for the management of both instance hardness and class imbalance independent of the ML algorithm in use. It adopts validation metrics suitable for imbalanced multi-class classification problems and includes several iterative phases that enable users to apply undersampling and oversampling in various strategic schemes.
Our contributions are summarized as follows:

\begin{itemize}
\item a coherent visual analytic workflow that takes into account instance hardness, while \hl{leveraging} undersampling and oversampling techniques;
\item a working prototype of the suggested workflow in the form of our VA system, \textsc{HardVis}, which comprises a novel combination of multiple coordinated views to support the entire process of selectively undersampling and oversampling parts of the data set;
\item a proof-of-concept showcasing the proposed system's applicability with a hypothetical usage scenario, and a use case \hl{that illustrates the utility of} our decision to deploy sampling approaches and involves humans in-between automated methods; and
\item the discussion of the methodology and findings of interview sessions with five ML experts, presenting positive results.
\end{itemize}

\noindent The remainder of this paper is organized as follows. \hl{In~\autoref{sec:relwork}, we review automated methods for the detection of different data types, visually-assisted identification of outliers and rare examples, and visualization approaches for data-centric ML error analysis.} Afterwards, in~\autoref{sec:goals}, we outline the analytical tasks and design goals for using VA to manage instance hardness in imbalanced data sets, and we emphasize the need for both automatic approaches and human intuition. \autoref{sec:system} presents the system's functionalities and simultaneously describes a first simple use case with multiple cycles of undersampling and oversampling applied to specific instances in order to enhance predictive performance. Following that, in~\autoref{sec:case}, we illustrate the applicability and utility of \textsc{HardVis} with two real-world data sets concentrating on detecting breast cancer and recognizing vehicles from their silhouettes. Thereafter in~\autoref{sec:eval}, we examine the input received from the expert interviews, including limitations identified by the experts. Subsequently, in~\autoref{sec:dis}, we reflect further on the visual design and the limitations of our work that lead to future plans for \textsc{HardVis}. Finally,~\autoref{sec:con} concludes our paper.

}

\section{Related Work} \label{sec:relwork}{%
  \hl{This section summarizes previous research on automatic approaches for the identification of different types of instances, visualization methods for outlier/anomaly and rare category detection, and data-centric ML solutions from the visualization community.} To underscore the uniqueness of our approach, we explain the difference between such solutions contrasted to \textsc{HardVis}. To the best of our knowledge, there is no literature explaining the use of VA for the complete undersampling and oversampling procedure, along with the partial application in specific types based on the visual exploration of data and distributions.

\subsection{Automatically Distinguishing Types of Instances} \label{sec:types}

In the ML community, several methods for automatically categorizing data instances into different types exist, with a particular focus on the outlier/anomaly detection research in the past decades~\cite{Chandola2009Anomaly,Hodge2004Survey}. Nevertheless, most algorithms cannot identify rare cases that are typically isolated groups, including a set of comparable data examples that deviate from the majority---rather than single isolated instances which are outliers. The majority of anomaly detection techniques can be divided into five categories: (1) classification-based~\cite{Hawkins2002Outlier,Wong2003Bayesian,Mahoney2003Learning}, (2) density-based~\cite{Bay2003Mining,Breunig2000LOF}, (3) clustering-based~\cite{Munz2007Traffic,Vatturi2009Category,Syarif2012Unsupervised}, (4) statistical-based~\cite{Yamanishi2004OnLine,Kwak2017Statistical}, and (5) ensemble approaches~\cite{Hulse2009Knowledge,Saez2015SMOTE,Vanerio2017Ensemble,Zhang2017LSHiForest}. The last category is a hybrid one, which aims to combine the benefits of the various techniques from the other categories. The problem with all the approaches, except for the density-based approaches, is the misalignment with sophisticated undersampling (e.g., NCR) and oversampling algorithms (e.g., ADASYN) that are using KNN to propose instances for removal or addition, respectively. Two empirical studies~\cite{Skryjomski2017Influence,Napierala2016Types} that were conducted with density-based sampling algorithms deploy KNN to distinguish the type of each instance along with multidimensional scaling (MDS)~\cite{Kruskal1964Multidimensional}, which is a global linear dimensionality reduction algorithm. \hl{We follow the same methodology to characterize instances based on local characteristics, but \textsc{HardVis} uses an interactive UMAP projection~\cite{McInnes2018UMAP} since it preserves mostly the local structure~\cite{Espadoto2021Toward}.} Although those studies suggest that applying sampling techniques in specific types of instances (e.g., by using only outliers) can boost predictive performance, controlling which subsets of particular instance types are considered when undersampling and oversampling is an undiscovered step. This research opportunity inspired us to design \textsc{HardVis}.

Density-based algorithms~\cite{Huang2011RADAR,He2008Graph} also work well with the detection of rare categories by discovering substantial changes in data densities using a KNN search in the high-dimensional space. But how to choose the best k-value for a given data set? While it is possible to estimate the best k-value automatically by using the local outlier factor~\cite{Breunig2000LOF}, the balance of the distribution of safe and unsafe instances could be off when focusing merely on rare cases and outliers. Huang et al.~\cite{Huang2014Rare} proposed a method for automatically selecting k-values. However, their algorithm starts with a seed depending on the target category, which is often difficult to set. iFRED and vFRED~\cite{Liu2014Prior} are two approaches for identifying rare categories based on wavelet transformation without the necessity of any predefined seed. Nevertheless, these methods are robust in low-dimensional data only but fail to discover the remaining types of data introduced in~\autoref{sec:intro}, which are important for \textsc{HardVis}. 
Regarding decision boundaries and borderline examples, Melnik~\cite{Melnik2002Decision} analyzes their structure using connectivity graphs~\cite{Martinez1994Topology}. And finally, Ramamurthy et al.~\cite{Ramamurthy2019Topological} utilize persistent homology inference to describe the ambiguity (or even lack) of decision boundaries. All described methods, while being valuable, do not focus on the problem of undersampling or oversampling at all, as it happens with our system.



\subsection{Visualization for Outlier and Rare Category Detection} \label{sec:detect}

Numerous VA approaches are combined with detection algorithms as described in~\autoref{sec:types}. Usually, they are designed for supporting outlier and rare categories identification and classification, which could be considered relevant to our work. Oui~\cite{Zhao2019Oui} is a tool that assists users in comprehending, interpreting, and selecting outliers identified by multiple algorithms. \#FluxFlow~\cite{Zhao2014FluxFlow} is another 
VA system that utilizes complex analytical methods to find, summarize, and understand aberrant information spreading patterns. TargetVue~\cite{Cao2016TargetVue} detects users with abnormal behaviors using the local outlier factor and intuitive behavior glyph designs. An extension of such glyphs, named as Z-Glyph~\cite{Cao2018Z}, was developed to aid human judgment in multivariate data outlier analysis. RCLens~\cite{Lin2018RCLens} is an active learning system that uses visualization approaches to support the discovery of rare instances. EnsembleLens~\cite{Xu2019EnsembleLens} is a hybrid visual system that utilizes a modified Gaussian mixture model~\cite{Arakawa2019REsCUE} to identify problematic patterns in human behaviors. RISSAD~\cite{Deng2021RISSAD} is an interactive approach that not only assists users in detecting abnormalities but also automatically defines them using descriptive rules. Even border detection has recently gotten some attention thanks to a VA method~\cite{Ma2021Visual} which uses the power of explainability from linear projections to help analysts study nonlinear separation structures. However, the final goal of \textsc{HardVis} differs since we try to merge the gap between instance hardness and sampling techniques for evaluating their suggestions. None of the above VA systems incorporate sampling mechanisms, as defined in~\autoref{sec:intro}.

VERONICA~\cite{Rostamzadeh2021VERONICA} is a domain-specific VA system that uses undersampling and SMOTE for specific classes of data and groups of features. Nonetheless, \textsc{HardVis} is inherently designed to be generalizable to any numerical data set stored in a tabular form. It also accounts for instance hardness while enabling the micromanagement of the sampling techniques. To improve the efficiency of model construction, Li et al.~\cite{Li2018Interactive} presented a VA approach that allows infusing dynamic user feedback in various forms, with interactive addition of new samples being one of them. Despite that, the goal is very different from ours since the focus is on learning with a limited amount of data or incrementally learning as in~\cite{Paiva2015An}. During the main use case of RuleMatrix~\cite{Ming2019RuleMatrix}, Ming et al. manually selected a problematic subset of instances and applied oversampling, which resulted in improved model accuracy. In contrast, \textsc{HardVis} enables knowledgeable users to systematically explore the distribution of data in different types and the suggestions of the undersampling and oversampling to enhance predictive performance.


\subsection{Data-centric Machine Learning}

\hl{Most of the model-centric ML work so far has focused on how model developers incrementally improve an existing or newly invented ML algorithm's predictive performance while making no changes to the collected data~\cite{Hamid2022From}. On the other hand, practitioners of data-centric ML maintain the ML model stable while iteratively upgrading the quality of the data at hand~\cite{Hamid2022From}. Advocates for data-centric ML have recently increased in volume. A few reasons for this shift are the benefits of involving domain experts in the data analysis process and the necessity for very configurable solutions that focus on subsets (or \emph{slices}) of data~\cite{Hamid2022From}. Closely related to this paradigm, ModelTracker~\cite{Amershi2015ModelTracker} and Squares~\cite{Ren2017Squares} are two interactive visualization approaches that improve a more standard confusion matrix to detect issues with particular instances and enable users to tune the input by monitoring the output of the model. The former proposes a visualization that incorporates information from a variety of typical descriptive statistics while providing instance-level performance and allowing for direct error analysis and troubleshooting. The latter computes performance measurements and assists users in concentrating their efforts on instance-level issues. Therefore, both works follow the general framework of visual parameter space analysis (vPSA)~\cite{Sedlmair2014Visual}. Although \textsc{HardVis} is also an applied example of the vPSA framework, it is explicitly designed for the first stages of an ML model-building pipeline, addressing a clear need for applying sampling techniques in specific types of instances only.}
 
\hl{Active learning is also part of data-centric ML solutions. It can be defined as the active usage of a learning algorithm to iteratively suggest to a user to classify unknown instances in order to increase the ML model's performance quickly~\cite{Settles2012Active}. In the visualization community, many VA techniques have been developed explicitly for active learning~\cite{Bernard2018Towards,Bernard2021A,Bernard2018Comparing,Grossmann2021Does}. More specifically, these works have focused on how VA can help users during the labeling process for semi-supervised learning problems. The challenges are somewhat similar to ours since understanding how hard (or important) it is for an instance to be labeled before the rest is a relatable problem. However, our end goal is to prioritize which instances should be undersampled and oversampled first (and how exactly) in supervised learning classification problems containing labels for all data instances.} 

\section{Analytical Tasks and Design Goals} \label{sec:goals}
  This section outlines the basic analytical tasks (\textbf{T1--T5}) that a user should be able to complete when undersampling or oversampling while using a VA system for support and direction. Following that, we present the design goals (\textbf{G1--G5}) that guided the development of \textsc{HardVis}.

\begin{figure*}[tb]
\centering
\includegraphics[width=\linewidth]{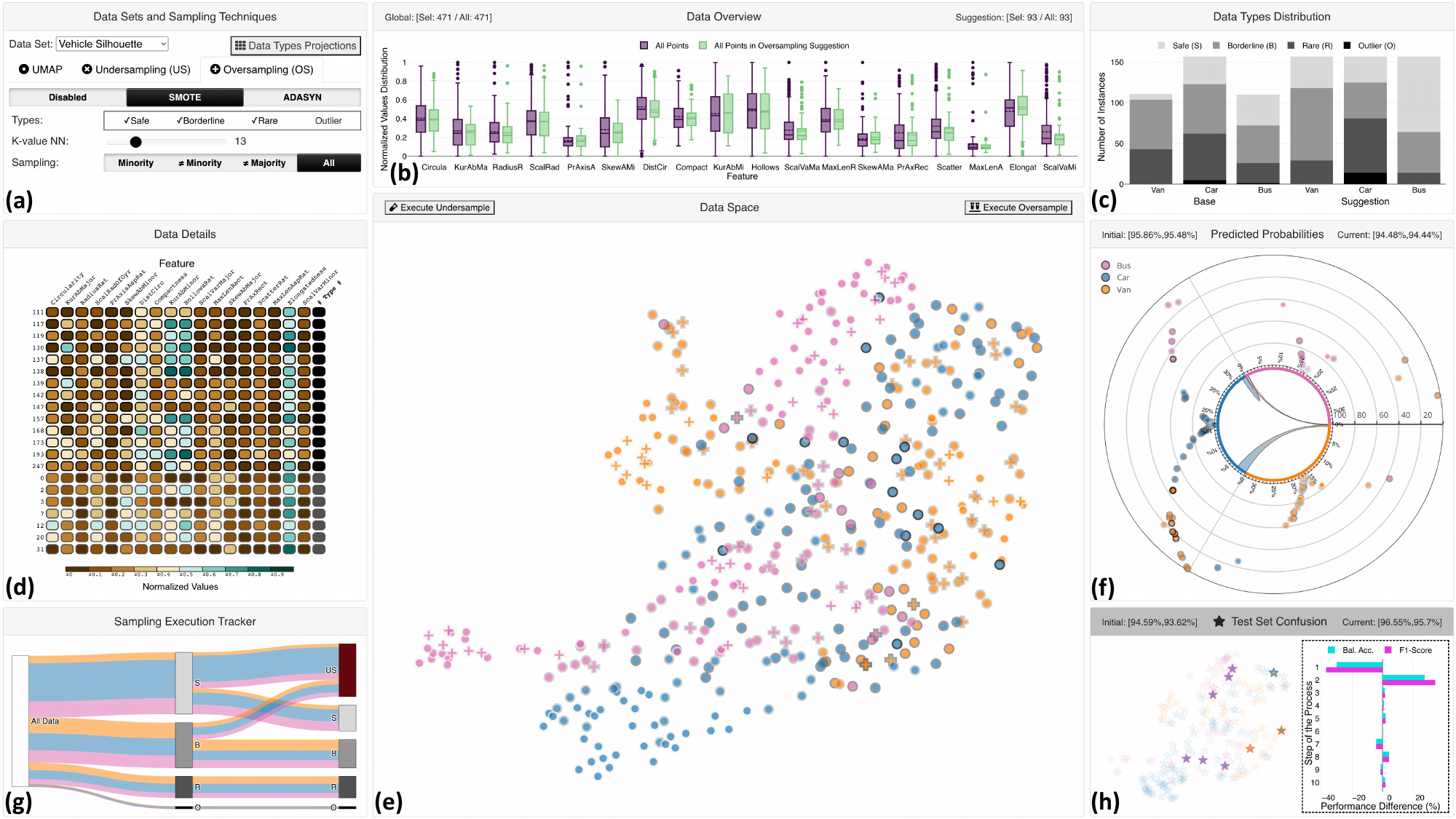}
\caption{Undersampling and oversampling certain data types with \textsc{HardVis}: (a) the panel with many tunable parameters for UMAP, undersampling, and oversampling; (b) box plots for comparing the values of all points against the algorithm's suggestion in each feature; (c) \hl{a} stacked bar chart showing the base vs. the new distribution if the suggestion is approved; (d) \hl{a} table heatmap view for comparing the instances' values across all features; (e) \hl{a} UMAP projection emphasizing the additions/deletions of points, along with the data type for every instance; (f) \hl{an} inverse polar chart with chords that depicts the predicted probabilities, as well as the training confusion; (g) \hl{a} Sankey diagram for tracking any undersampling or oversampling confirmed actions; and (h) \hl{a} visual embedding based on (e) to highlight the confusing test instances, and \hl{a} horizontal bar chart to illustrate the performance difference for each step.}
\label{fig:teaser}
\end{figure*}

\subsection{Analytical Tasks for Undersampling and Oversampling}

From the in-depth examination of the related work highlighted in~\autoref{sec:relwork} and our own recent experiences implementing VA tools for ML~\cite{Chatzimparmpas2020t,Chatzimparmpas2021StackGenVis,Chatzimparmpas2021Visevol,Chatzimparmpas2022FeatureEnVi,Chatzimparmpas2022VisRuler}, we came up with five analytical tasks. 

\textbf{T1: Identify the various types of instances.}
As the decrease in predictive performance is connected to data distribution-related factors, such as the presence of many rare subgroups obscuring the classification~\cite{Weiss2000Quantitative,Japkowicz2001Concept}, the consequences from the overlap between the classes~\cite{Prati2004Class,Garcia2007An}, or the existence of several misclassified examples~\cite{Napierala2010Learning}, a primary goal is to spot such groups of points---as precisely as possible---with the use of VA systems.

\hl{\textbf{T2: Support the exploration of undersampling vs. oversampling alternatives applied globally and locally.}}
When applying such techniques, the data instances used as input for undersampling and oversampling algorithms could differ depending on the stable anchors a user sets. An example of a stable anchor is how the partitioning of data into four types occurs, leading up to 16 different SBRO combinations used as input for the sampling algorithms. Also, the distribution of SBRO (as defined in~\textbf{T1}) is another factor to be considered as a stable anchor under investigation.
On the one hand, global undersampling or oversampling will allow all instances to be candidates for removal or under consideration when creating synthetic data, respectively. On the other hand, locally applied algorithms will dynamically enable users to consider local characteristics of data points and exclude a few suggestions from the pool of recommendations. Modifying this ratio dynamically could be beneficial for the ML model, thus the user's interaction guided by visual feedback is necessary.

\textbf{T3: Explore automated methods' suggestions.}
The identification of conditions for the efficient use of a particular method is an open research problem~\cite{Napierala2016Types}. A user should be competent in judging the influence of a suggestion on the whole data set. For example, what if, by removing too many rare cases, the model overfits the training data but generalizes poorly in a test set? A user should be empowered by VA systems that facilitate exploratory analysis of unsafe instances.

\textbf{T4: Confirm suggestions by making justifiable decisions.} 
A user should have the ability to partially confirm the proposal of the automated methods based on the analysis he/she has performed earlier in the preceding task. How will the data distribution change due to the acceptance of such a suggestion? VA systems should envision these future steps and enhance users' decision-making.

\textbf{T5: Monitor and evaluate the results of the sampling process.}
At any stage of the sampling process (\textbf{T2--T4}), a user should be able to observe performance fluctuations with the use of appropriate validation metrics for imbalanced data sets (e.g., \emph{balanced accuracy} and \emph{f1-score}). A user might also wish to look back at the history of activities to see if any crucial actions corresponded to better results. Thus, VA systems must be capable of providing ways to monitor performance.

\subsection{Design Goals for \textsc{HardVis}}

\hl{We identified five design goals for our system to meet in order to fulfill the more general aforementioned analytical tasks for undersampling and oversampling. We implemented them in~\autoref{sec:system}.}

\textbf{G1: Visual examination of several data types' distributions and projections to choose a generic `number of neighbors' parameter.} 
Our goal is to assist in the search for distinctive distributions of data types that might consider different populations of SBRO instances (\textbf{T1}). By systematically modifying the \emph{number of neighbors} parameter of UMAP, we aim to assure that users will pick a better value based on the visual exploration of data types in the generated projection. Furthermore, this value propagates in the undersampling and oversampling techniques that require a \emph{k-value}, which works similarly to the above parameter. 

\textbf{G2: Application of undersampling and oversampling in specific data types only, with different parameter settings.}
There are several different undersampling and oversampling techniques, but they are usually only applied the entire training set \hl{(i.e., global sampling)}. However, with our proposed system, we enable users to choose a technique, tune the parameters depending on the visual exploration, and even deploy them in particular subsets of the training data (\hl{i.e., local sampling as established in \textbf{T2}}).

\textbf{G3: Exploratory data analysis of unsafe suggestions.}
Next, the system should provide sufficient visual guidance to users to focus on the exploration of the values in each feature for unsafe suggestions (\textbf{T3}). The analysis of borderline, rare, and outlier data types should be feasible in a generic and detailed manner.

\textbf{G4: Comparison of trade-offs while removing or adding training instances throughout the decision-making process.}
After the extraction of evidence as defined in \textbf{G3}, users should see how the distribution of instances will change due to the undersampling and/or oversampling phases. Next, the system should give a prediction for a data point and juxtapose it to all other points. With this, users should be able to estimate the impact of algorithmic recommendations during exclusion or inclusion of instances (\textbf{T4}).

\textbf{G5: Keep track of critical steps and evaluate predictive performance in general and for specific test instances.} Users' interactions should be tracked in order to preserve a history of modifications in the training set, and the performance should be monitored with validation metrics (\textbf{T5}). Finally, using an unseen test set, the system should continuously stress the difference in the model's predictive performance.

\section{HardVis: System Overview and First Application} \label{sec:system} {%
  \begin{figure}[tb]
\centering
\includegraphics[width=\linewidth]{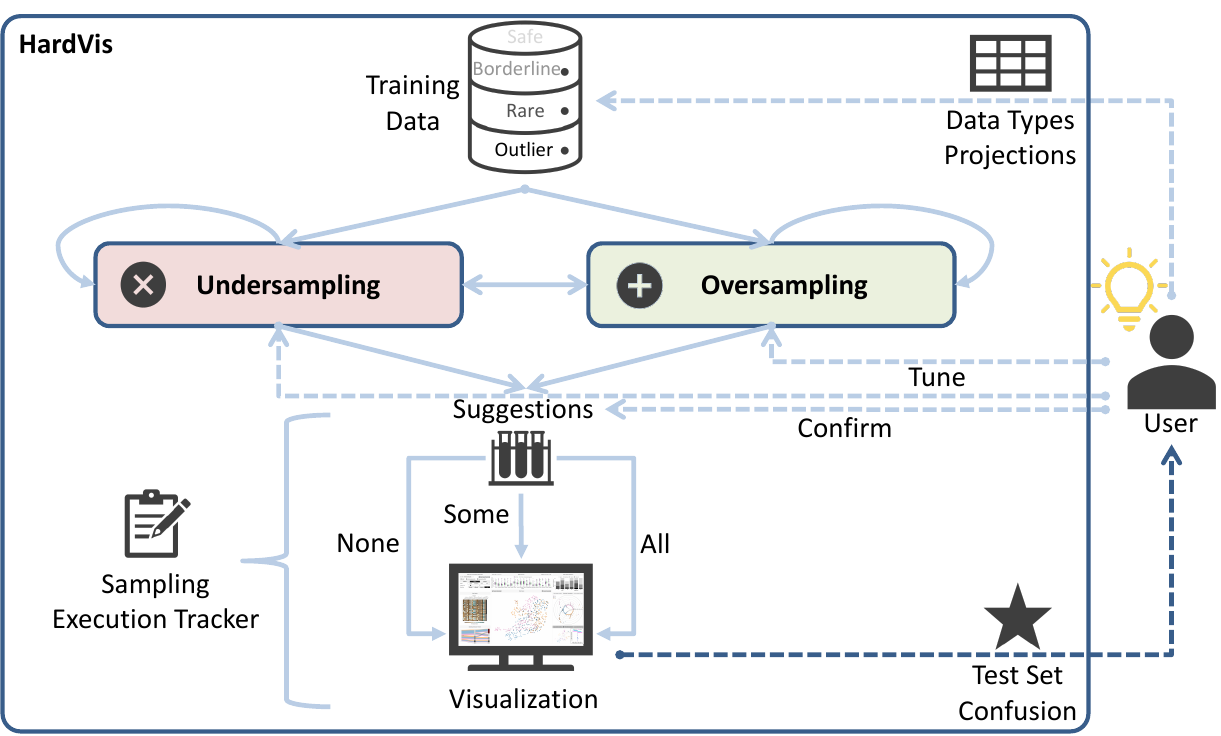}
\caption{The \textsc{HardVis} workflow starts by classifying the training data into four types according to the user's visual inspection of 9 alternative projections. The data is sent for either undersampling or oversampling, which can make suggestions continuously. The user's confirmation is requested after the exploratory data analysis through the visualizations.}
\label{fig:workflow-diagram}
\end{figure}

Following the analytical tasks and the resulting design goals, we have developed \textsc{HardVis}, an interactive web-based VA system that allows users to identify areas where instance hardness occurs and to micromanage sampling algorithms. \hl{\autoref{sec:addlimit} contains further implementation details.}

The system consists of 8 interactive visualization panels (\autoref{fig:teaser}): (a) data types projections ($\rightarrow$  \textbf{G1}) incl. data sets and sampling techniques ($\rightarrow$  \textbf{G2}), (b) data overview, (c) data types distribution, (d) data details, (e) data space, (f) predicted probabilities ($\rightarrow$  \textbf{G3} and \textbf{G4}), (g) sampling execution tracker, and (h) test set confusion ($\rightarrow$  \textbf{G5}). 
We propose the following \textbf{workflow} for the integrated use of these panels (cf. \autoref{fig:workflow-diagram}): 
(i) explore various projections with alternative distributions of data types, leading to the division of training data into SBRO (cf. \autoref{fig:iris-under}(b));
(ii) in the undersampling or oversampling phase, tune the active algorithm's parameters to affect specific types of data (\autoref{fig:teaser}(a));
(iii) during the confirmation phase, identify which suggestions will impact negatively or positively the predictive performance and approve or reject any suggestion (cf. \autoref{fig:teaser}(b)--(f)); and
(iv) store every manually-operated sampling execution, identify confused test instances, and compare the predictive performance in each step of the process according to two validation metrics designed explicitly for imbalanced classification problems (\autoref{fig:teaser}(g) and (h)).
These steps are iterative, and they might occur in any sequence. The created knowledge obtained from the undersampled/oversampled data set is the end result. This knowledge can be useful to users that have to explain and are accountable for their actions, e.g., people working in critical domains such as medicine. 

\begin{figure*}[tb]
  \centering
  \includegraphics[width=\linewidth]{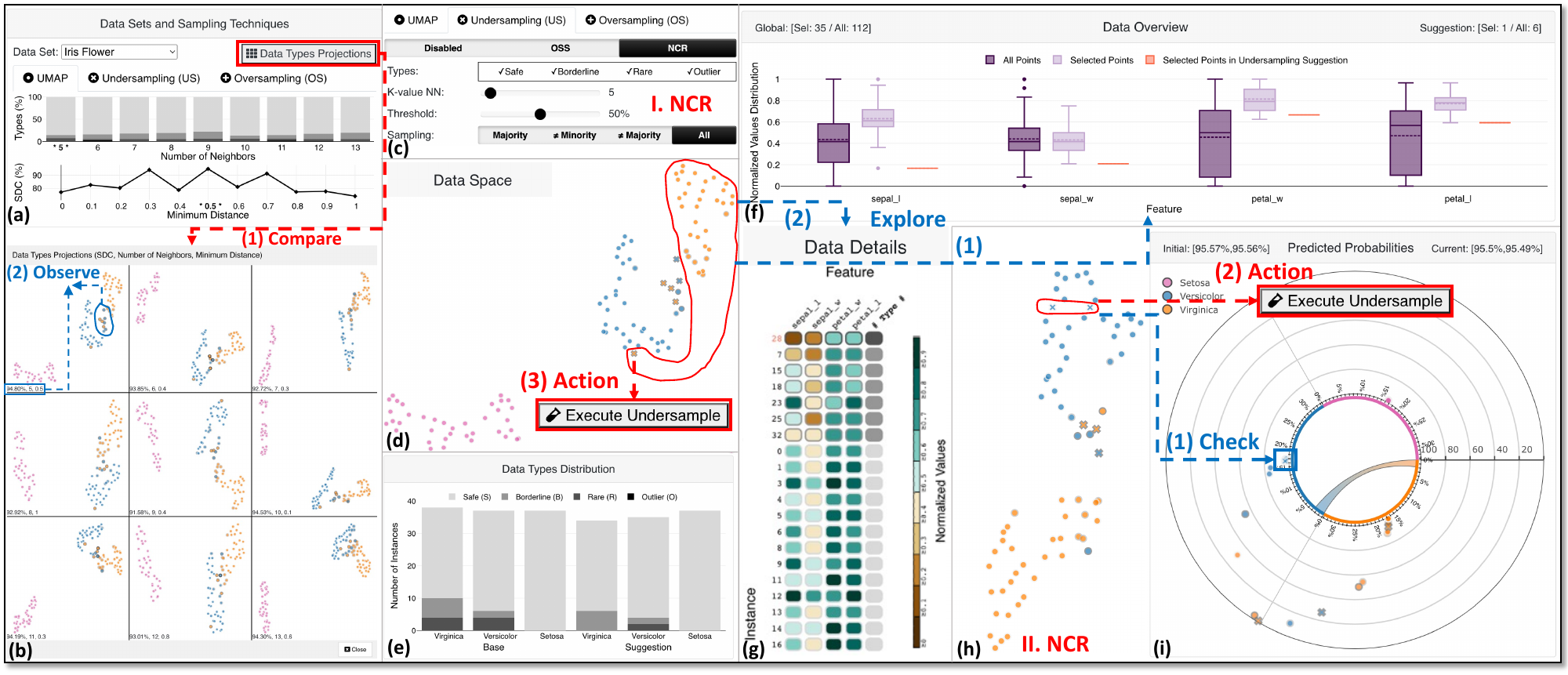}\vspace{-1mm}
  \caption{At first, a comparison of different data types projections and then two consecutive undersampling phases with the NCR algorithm are shown in this arrangement of screenshots. The \emph{default} value for the number of neighbors is 5 (see (a)), which is used as input for computing the type of each instance with KNN. The projections are generated by systematically tweaking the above parameter, as illustrated in (b); the best choice is theoretically the highest value for the Shepard diagram correlation (SDC) metric.  In (c), we have activated the algorithm, and we check the impact of this automated technique on the projection in (d). (e) presents the difference in distributions of all data types per class label from when the algorithm was inactive as opposed to its activation. In (f), we explore a specific rare case under removal consideration. This instance is contrasted against the remaining points of this same class (i.e., virginica in orange color); the selection was made using a lasso interaction, as demonstrated in (d). While the values for all features are lower for this sample than the rest, \emph{sepal\_l} appears the furthest away. Additional details can be found in (g) that highlights these differences in values of particular features and confirms our findings from the data overview. Consequently, we choose to delete this instance because it might cause further confusion to the model, as depicted in (d). The second time we deploy NCR (cf. (h)), two safe instances are in our focus since they are easily classified due to the high predicted probability visible from the inverse polar chart in (i). Therefore, we decide to remove these two points.} \vspace{-2mm}
  \label{fig:iris-under}
\end{figure*}

\textsc{HardVis} employs a cutting-edge ensemble learning approach named as XGBoost~\cite{Chen2016XGBoost}, and its workflow is model-agnostic. To make our approach even more future-proof, we train this ML algorithm with the \emph{Bayesian Optimization} package~\cite{Bayesian}. \textsc{HardVis} utilizes OSS, NCR, SMOTE, and ADASYN, which are state-of-the-art sampling algorithms that are tweaked to receive specific SBRO instances as an input. Despite that, these algorithms are easily replaceable. The reader is referred to~\cite{He2009Learning,He2013Imbalanced} for a more detailed analysis of different strategies that cope with class imbalance. For this section and the use cases in~\autoref{sec:case}, we split the data sets into 75\% training and 25\% testing sets with the stratified strategy (i.e., keeping the same balance in all classes for both sets) and validate our results with 5-fold cross-validation. Also, we scan the hyperparameter space for 25 iterations, choosing the model with the best accuracy. The hyperparameters we used are the same as in another VA system developed by us~\cite{Chatzimparmpas2022FeatureEnVi}.

In the following subsections, we explain the system by using a running example with the \emph{iris flower} data set~\cite{Fisher1936The} obtained from the UCI ML repository~\cite{Dua2017}. The data set represents a balanced multi-class classification problem and consists of four numerical features and 150 instances. The three classes are: \emph{setosa}, \emph{versicolor}, and \emph{virginica}.

\subsection{Data Types} \label{sec:data}

\textsc{HardVis} follows the Napierala and Stefanowski~\cite{Napierala2016Types} methodology in order to label all training instances in one of the following types: safe (S) examples, borderline (B) samples, rare (R) cases, and outliers (O). To calculate the difference between instances in the high-dimensional space, we use KNN~\cite{Altman1992Introduction,Fix1989Discriminatory} with the \emph{default} value of k being 5 and the \emph{euclidean} distance metric. For determining the type of a sample with $\mbox{k}=5$, we would have, e.g., 5 or 4 nearest instances being from the same class, then the sample gets labeled as S; 3 or 2 instances from the same class, then it belongs to B; only 1 instance from the same class, it is R; and 0 (i.e., the 5 nearest instances are from the other class), it becomes O. However, the analogies will change with $\mbox{k}>5$.

As shown in~\autoref{fig:iris-under}(a), stacked bar chart, the distributions of instances change accordingly as the \emph{number of neighbors} in the UMAP~\cite{McInnes2018UMAP} shifts since we utilize the same value for the KNN algorithm. \hl{Thus, the goal of the two-dimensional projection is to reflect visually the same separation of training instances into the SBRO types.} The \emph{minimum distance} is another parameter of UMAP that (in our case) is being automatically computed from the maximum achievable Shepard diagram correlation (SDC)~\cite{Chatzimparmpas2020t} score (see \autoref{fig:iris-under}(a), line chart). This metric serves as a first indicator of optimal distance preservation between the low- and the high-dimensional space. Nevertheless, it cannot be trusted blindly, and human exploration is necessary to conclude which parameters are optimal for the given data set.

The main challenge of KNN is the user-selected k-value, thus it is a highly parametric-dependent approach. To resolve this problem, we enable the user to explore different data types' projections generated by the systematic change of k-value from 5 to 13 (cf.~\autoref{fig:iris-under}(b)). This range is chosen intentionally because, in low k-values, a slight modification is more impactful to the projection~\cite{Napierala2016Types}. \hl{However, these values are adjustable within the code.}

\subsection{Undersampling} \label{sec:under}

\autoref{fig:iris-under}(c) presents the tab for Undersampling (US), which along with the standard method's parameters comprises a Types menu with options to exclude any SBRO group. The k-value is automatically tuned due to the selection of the number of neighbors parameter, as explained in~\autoref{sec:data}. OSS~\cite{Kubat97Addressing} uses Tomek links~\cite{Tomek1976An} which are ambiguous points on the class boundary that are typically identified and removed. Moreover, it employs the condensed nearest neighbor rule~\cite{Hart1968The} to remove redundant examples far from the decision boundary. In contrast, NCR~\cite{Laurikkala2001Improving} is an undersampling technique that combines the condensed nearest neighbor rule to exclude redundant examples and the edited nearest neighbors rule~\cite{Wilson1972Asymptotic} to remove noisy or ambiguous points. Its main difference from OSS is that fewer redundant examples are deleted, and more attention is placed on ``cleaning'' those retained instances. Each algorithm expects input for a unique parameter. In particular, NCR has \emph{Threshold} which is used for deciding whether to consider a class or not during the cleaning after applying edited nearest neighbors. \emph{Seeds} is the number of samples to extract in order to build a set S for OSS. All these techniques can be employed in the Majority, $\neq$~Minority, $\neq$~Majority, or All classes according to the user's choice. In multi-class classification problems, the Majority will be merely the class that contains the most instances, $\neq$~Minority will be all classes except the one with the least instances, and so on. In balanced data sets, only the All option is relevant.

The UMAP projection in~\autoref{fig:iris-under}(d) allows users to observe the type of each instance concurrently and if it was suggested for removal with an ``$\times$'' symbol or addition with a ``$+$'' mark by the active undersampling or oversampling algorithm, respectively. The parameters for the UMAP are set as discussed in the preceding subsection. Hovering over a point will present details on demand such as the ID of the point, the predicted probability, and the values for each feature. 

The distribution of data types is known due to a stacked and grouped bar chart with the instances distributed in SBRO and per class, simultaneously (cf.~\autoref{fig:iris-under}(e)). The base distribution is also comparable with the suggestion from the sampling algorithm that will modify the initial distribution.

\autoref{fig:iris-under}(f) is a box plot that facilitates the comparison of all points per feature versus the selected points via lasso functionality in the projection. When a sampling algorithm is active, the same group of instances with merely the sampling suggestions is also visualized. In case of no selection, a simpler version of all points against all points in either undersampling or oversampling suggestion exists (see~\autoref{fig:teaser}(b)). Users' actions determine the mode automatically. The features are sorted from left to right, from the least important to the most important at each execution step of undersampling/oversampling (due to XGBoost retraining process). The proposals for removal are denoted in light red color, and light green is used for the suggested additions. 

The table heatmap view in~\autoref{fig:iris-under}(g) is a more detailed view of the aggregated results present in the box plots. It normalizes the values from 0 to 1, evident in dark brown to dark teal colors, and it shows for each feature the current value in each instance. The features are sorted as in the box plots. Moreover, the \emph{\# Type \#} is perceivable through this visual representation, with outliers, then rare cases, next borderline examples, and finally safe instances being at the top of the list. The selection of a specific feature in this view applies the diverging colormap to the projection for comparing all instances for this particular feature (see~\autoref{fig:vehicle}(d), Zoomed in View). More detailed discussions on the visual design behind some of the views can be found in~\autoref{sec:design}.

The inverse polar chart in~\autoref{fig:iris-under}(i), is deliberately designed to provide more space to instances that are in the borders between two classes or completely misclassified cases. The predicted probability with the ground truth class is used for the 100 to 0 axis, and the angle/orientation is computed as the difference in predicted probability of belonging to the remaining two classes. The greater this difference is, the farther a point deviates from the center of its circular segment corresponding to the correct class label. In our example, the versicolor has a few instances mainly confused with the virginica and vice versa. This is why all setosa instances are near 100\% predicted probability in the purple circular segment. 
The size of each piece is calculated from the number of training instances that belong to a particular class, with extra space being provided to larger classes (i.e., consisting of more points). The same symbols as in the projection are also retained here. This approach can easily work for two or three classes but becomes challenging to interpret with more classes; such limitations are discussed in Sections~\ref{sec:eval} and~\ref{sec:addlimit}. The centerpiece of this visualization is a chord diagram that summarizes the confusion matrix for the training data, as in~\cite{Alsallakh2014Visual}. Thus, in~\autoref{fig:iris-under}(i), the confusion between versicolor and virginica is immediately distinguishable by the chords linking the different circular segments. The number of confused instances from one to the other classes is encoded as chord width.

\subsection{Oversampling} \label{sec:over}

Two mainstream oversampling techniques are implemented in \textsc{HardVis}. SMOTE~\cite{Chawla2002SMOTE} finds the KNN in the minority class for each of the samples in the class. Next, it draws a line between the neighbors and generates random points on the lines. ADASYN~\cite{He2008ADASYN} is the same as SMOTE, just with a minor improvement. After creating those samples, it adds a small random deviation to the points, thus making it more realistic with the additional variance. Similar to the undersampling techniques before, SMOTE and ADASYN have all options except for the division of types provided via a separate menu of our system. The All option is equivalent to $\neq$ Majority, but we implemented them differently when a type of instance is deactivated. The former considers removing all points of the specific deactivated type/s irrelevant to the class that will be oversampled, leading to more excluded points for the active algorithm. The latter excludes from the pool of points only those from the deactivated type/s but from the class that will be oversampled. The Minority and $\neq$ Minority are implemented based on the second schema described here. The same exploration and analysis options mentioned in the previous subsection also apply for oversampling.

 \begin{figure*}[tb]
  \centering
  \includegraphics[width=\linewidth]{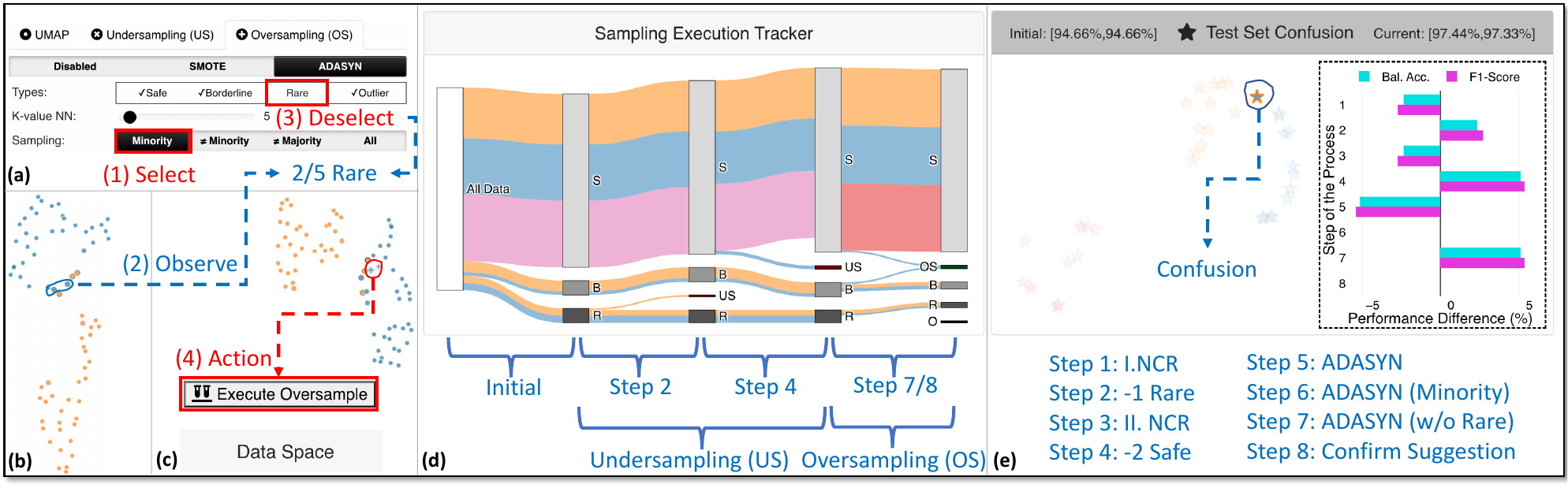}\vspace{-1mm}
  \caption{An oversampling phase that aims to balance the data set again. According to (a), we use ADASYN for the minority class (versicolor in blue) that contains fewer instances. Also, we exclude from the input of the algorithm two rare cases near the borders of two classes, as illustrated in (b). The system proposes two artificially-created points for addition that we approve, see (c). The Sankey diagram in (d) summarizes the core execution phases with undersampling and oversampling steps. Only one test instance is confused according to (e), while the manual decisions (step 2, step 4, and step 7/8) improved the balanced accuracy and f1-score scores compared to the automated methods (steps 1, 3, and 5).}
  \label{fig:iris-over}
\end{figure*}

\subsection{Sampling Execution Tracker and Test Set Confusion} \label{sec:results}

Each manual undersampling or oversampling confirmation is registered in the Sankey diagram (see \autoref{fig:iris-over}(d)). The initial setting is to record the distribution of all training data to the SBRO types. Then, as an undersample or oversample execution takes place, the instances move from their type to the US (in dark red) or OS (in dark green) bin of the Sankey diagram. 

The test set is also plotted using the visual embedding of training data in each step (cf. \autoref{fig:iris-over}(e), left). All test instances are transparent when predicted correctly by the ML model and opaque in cases of confusion. For example, in~\autoref{fig:iris-over}(e), left, the star with blue color \hl{is from the versicolor class}, but it was predicted as virginica due to the orange outline. 
Furthermore, the initial and current balanced accuracy (bright turquoise) and f1-scores (deep magenta) are visible in the text at each side of the heading of the Test Set Confusion panel. The difference in performance based on those metrics is tracked for every step of the process with a horizontal bar chart (\autoref{fig:iris-over}(e), right).

\subsection{First Application} \label{sec:application}

In our first application, we observe that the maximum SDC value is 94.80\% (high correlation, \autoref{fig:iris-under}(b)), resulting in a most probably trustworthy projection. Another reassurance stems from the visual inspection of points in the middle of two classes that appear clearly confused, with most rare and borderline instances being located there.

The undersampling phase is perhaps most crucial since removing unsafe instances without justifying one's action could cause a severe issue to the ML model. We choose to activate the de-facto NCR algorithm without any tweaks to check the suggestions (\autoref{fig:iris-under}(c) and (d)). The distribution of instances changes according to this global suggestion for removal of orange and blue points, as seen in~\autoref{fig:iris-under}(e). Despite that, we want to explore further a suggestion for removal that is a distant point from the core virginica cluster. We use the lasso to select those points and proceed with the investigation. The box plot in~\autoref{fig:iris-under}(f) enables us to conclude that this is an extreme case relatively different from the remaining selected points of its own class since the values for all features are very low. The table heatmap view in~\autoref{fig:iris-under}(g) reaffirms our hypothesis, because the instance with ID 28 has the lowest \emph{sepal\_l} value ($<$ 0.1 due to dark brown color). We exclude this instance, but we keep the rare cases around the borders of the two classes that can easily flip class labels. Another phase of undersampling is also capable with \textsc{HardVis} since the new data become the ground zero for the next application of the automatic algorithm; NCR is again our choice. This time, five instances are proposed for deletion (cf.~\autoref{fig:iris-under}(h)). However, by checking the inverse polar chart in~\autoref{fig:iris-under}(i), we see that two of them are easily predictable and potentially redundant for the ML model. Therefore, we decide to exclude those two safe samples solely.

Using the Oversampling (OS) tab, we try to balance the classes that contain fewer training samples. In~\autoref{fig:iris-over}(a), we activate ADASYN for the minority class, which requires two more examples to restore balance in the training set. This setting, in combination with the observation of two rare cases that are in the borders of the versicolor class (\autoref{fig:iris-over}(b)), leads to the deselection of the Rare type. Consequently, these two rare cases are excluded from the pool of available for oversampling training instances. Without the appropriate choice of k-value, resulting in an expressive and effective distribution of data types, it would have been challenging to detect and handle such cases (especially if no class labels were provided).
The oversampling generated two instances that we accept in step 8, as depicted in~\autoref{fig:iris-over}(c).

\begin{figure*}[tb]
  \centering
  \includegraphics[width=\linewidth]{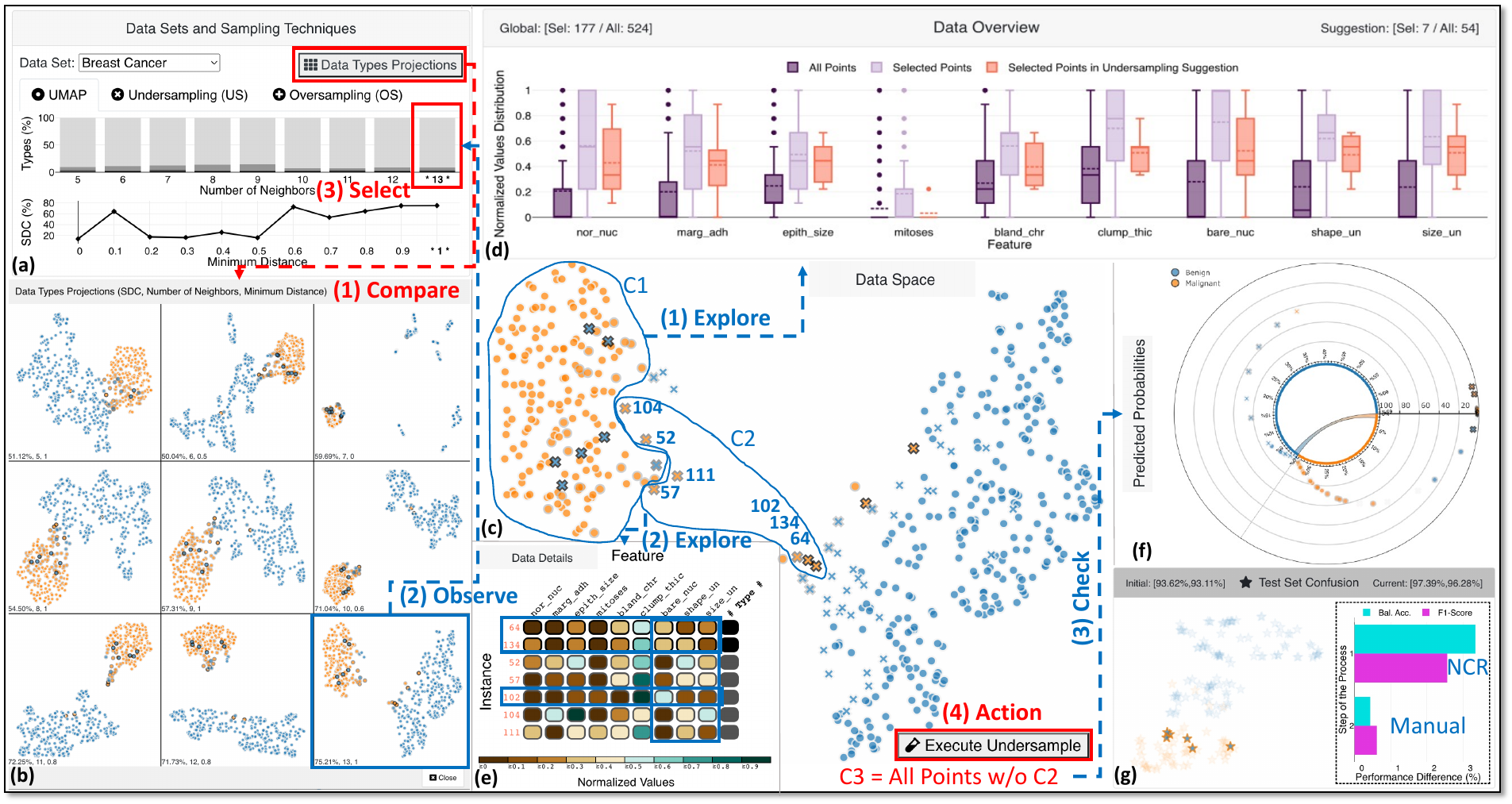}\vspace{-1mm}
  \caption{The investigation of diverse structures of data types and alternative suggestions in an undersampling scenario. View (a) shows the selection of the number of neighbors value of 13, which has 75.21\% Shepard diagram correlation (SDC) score, as illustrated in (b). The UMAP visible in (c) has one rare sample and 6 outliers belonging to the benign class that holds relatively normal values compared to the malignant cluster (C), as shown in (d). Therefore the suggestions for removal in C1 are valid since even humans cannot understand why these points are benign cases. On the other hand, C2 contains five rare examples and two outliers that serve as a bridge between the two classes, cf. (c). Interestingly, the three most important features differentiate the right group of points (IDs: 64, 102, and 134) from the left, i.e., \emph{size\_un}, \emph{shape\_un}, and \emph{bare\_nuc} in (e). This diversity is crucial when predicting difficult to classify instances, hence the analyst chooses to keep this cluster despite the NCR algorithm's suggestion for removal. C3 is the final selection, with most outliers being removed because the model badly predicted them, as seen in (f). This leads to (g), which presents an improved performance with 6 confused test instances that are cancer-free but predicted as the opposite. The malignant class is secure due to the rare cases being intact.}
  \label{fig:breast} \vspace{-1mm}
\end{figure*}

In~\autoref{fig:iris-over}(d), the deletion of one rare instance during the first NCR phase, the removal of two safe instances during the second NCR phase, and the oversampling phase utilizing ADASYN for generating a safe and a borderline instance is visible. Also, the confusion of a test instance is highlighted in~\autoref{fig:iris-over}(e) with the decisions of the automatic algorithm hurting the performance and the manual decisions in steps 2, 4, and 7/8 improving the predictive power of the ML model.
 
\section{Use Cases} \label{sec:case} {%
  In this section, we present a hypothetical usage scenario and a use case about how \textsc{HardVis} can evaluate suggestions based on local data characteristics to build trust in ML \hl{and to improve the balanced accuracy and f1-score scores for both training and testing sets.}

\subsection{Usage Scenario: Local Assessment of Undersampling}
Supposedly Zoe is a data analyst in a hospital, working primarily with healthcare data. She receives a manually-labeled data set with 9 features related to \emph{breast cancer}~\cite{Dua2017Machine}. This data set is rather imbalanced, with 458 \emph{benign} and 241 \emph{malignant} cases. From her experience, she knows that instance hardness and class imbalance can be troublesome for the ML model. Thus, she wants to experiment with well-known algorithms for undersampling and oversampling the data. However, especially with medical records, the use of merely automated methods is questionable because they cannot be trusted blindly. The doctors need explanations, and the minority class in this binary classification problem is of more importance than the majority consisting of healthy patients. \hl{In reality, patients who are healthy but predicted as ill will undergo extensive follow-up diagnostic tests before treatments such as surgery and chemotherapy are advised; however, the opposite is not true. To accomplish this main objective and to control the sampling techniques, Zoe deploys \textsc{HardVis}.}

\textbf{Choosing an accurate projection.}
Zoe begins with the selection of a \emph{number of neighbors} parameter by activating a window containing data types projections (cf.~\autoref{fig:breast}(a)). \textsc{HardVis} enables her to compare a grid of diverse projections, as presented in~\autoref{fig:breast}(b). The one with the highest SDC score (i.e., 75.21\%) is a noteworthy candidate because the two classes are clearly separated. Rare cases and outliers are also easily visible, forming a bridge between benign and malignant instances. She clicks on the bar with the number 13 in~\autoref{fig:breast}(a), and this projection becomes the main for further exploration. \hl{At this initial phase, 6 benign test instances were incorrectly classified, while the remaining 4 out of the 10 misclassified patients were actually malignant cases.}

\begin{figure*}[tb]
  \centering
  \includegraphics[width=\linewidth]{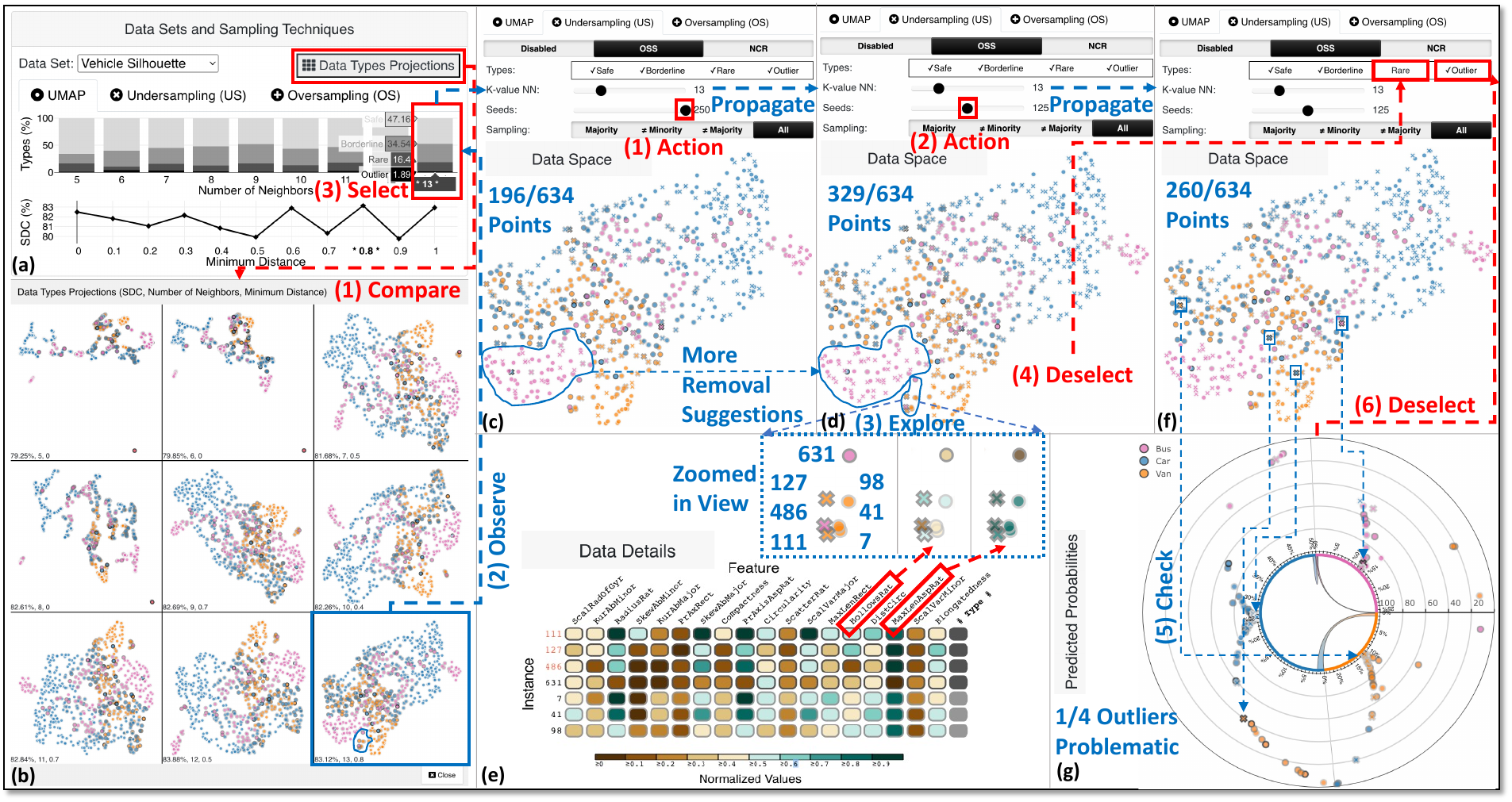}\vspace{-1mm}
  \caption{The examination of diverse structures of data types and alternative suggestions while undersampling with OSS. View (a) shows the selection of the number of neighbors value of 13, which is also used as input to KNN for sampling similarly in the high-dimensional space. This decision was made after a careful review of the 9 projections in (b), leading to a distribution of mostly safe instances, then borderline examples, next rare cases, and finally a few outliers. This is the second-best projection in terms of Shepard diagram correlation (SDC) score, but it preserves exceptionally well the clusters of buses in purple vs. vans in orange color at the bottom-left region. In (c), we experiment with the maximum seed value for the OSS algorithm, but it seems that several safe instances that could have been proposed for removal are actually not. Thus, we reduce the seed value in half to check if the new suggestion fits our viewpoint, as depicted in (d). The goal we set has been accomplished, however, a cluster of rare cases mixed in two classes is about to be deleted. In (e), we investigate further this group of points at the bottom-left corner; these instances differ mainly because of either \emph{MaxLensAspRat} or \emph{HollowsRat}. Such rare cases are critical information for the model to split these two classes; hence, we exclude the rare cases from the automatic algorithm, see (f). From the remaining 260 under consideration instances, there are four outliers that caught our interest. Only 1 out of the 4 outliers appears problematic based on (g). As a result, we exclude the outliers from the analysis and Execute Undersample to the remaining 256 points.}
  \label{fig:vehicle} 
\end{figure*}

\textbf{Examining unsafe instances proposed for removal.} 
Afterwards, she activates the NCR algorithm with the default settings (k-value is synchronized to 13 due to the previously-selected projection) from the Undersampling (US) tab. Cluster 1 (C1) in~\autoref{fig:breast}(c) is interesting because 7 benign cases \hl{(mostly marked as outliers)} are in between the malignant class. She chooses to compare the selected points in C1 against these suggestions of undersampling, as depicted in the box plots (\autoref{fig:breast}(d)). In summary, the values are lower for these points but still in between normal margins. Therefore, it would have been almost impossible for the doctors to conclude that these are healthy patients with benign cancer. \hl{A thorough check should be performed in these cases, e.g., to determine if the labels are erroneous. She first notifies the data collection team and doctors about this important finding and then removes C1 suggestions.} On the contrary, C2 includes five rare cases and two outliers with \emph{size\_un}, \emph{shape\_un}, and \emph{bare\_nuc} features separating the points closer to the benign class from the rest, as illustrated in the table heatmap view (\autoref{fig:breast}(e)). The right group of points has mostly lower values for the \emph{size\_un} and \emph{shape\_un} features, while the \emph{bare\_nuc} is higher compared to the points on the left. Zoe understands that such diversity is important when dividing borderline patients located at the conjunction of the two huge clusters. Therefore, she uses lasso selection to grab all points except for C2, which will be her manual undersample strategy. The inverse polar chart in~\autoref{fig:breast}(f) highlights the training instances that will be deleted, which are mostly completely misclassified instances or safe examples. The samples between the two classes already explored remain intact, which is essential since they all belong to the more important minority class. In~\autoref{fig:breast}(g), Zoe observes that only 6 test instances were incorrectly classified as having malignant cancer while they were healthy. When inspecting the Balanced Accuracy and F1-score scores, the overall predictive performance \hl{for the test set} seems slightly improved contrasted to the automatic algorithm. 
\hl{Nevertheless, the major gain is that the doctors might trust this modified data set more because the model correctly predicts all patients with malignant cancer (since there is no highlighted yellow star for the test set in~\autoref{fig:breast}(g), left).} Based on prior findings~\cite{Napierala2016Types}, Zoe stops her exploration at this phase because oversampling is ineffective for data sets with mostly safe instances.

\subsection{Use Case: Explorative Sampling for Better Classification}
This use case is about a multi-class classification problem. There are 18 features collected for the \emph{vehicle silhouettes} data set~\cite{Siebert1987Vehicle}. 
With the main task of classifying 199 \emph{vans}, 218 \emph{buses}, and 429 \emph{cars}, the class distribution is somewhat unbalanced. 

\textbf{Comparing projections and distributions of data types.} 
Similar to the procedure described in \autoref{sec:system}, we start by exploring which projection represents the data types in the best possible way. Three projections reaching high enough SDC with \emph{minimum distance} parameter equals to zero are extremely condensed, making it hard to observe anything (see \autoref{fig:vehicle}(b)). Among the remaining, two of them have SDC score of more than 83\%. Although they are two similar projections, the last one clearly shows the difference between bus and van classes in purple and orange, respectively (see the circled area at the bottom). We choose to continue with this projection; thus, we go back to \autoref{fig:vehicle}(a) and select a \emph{number of neighbors} parameter equal to 13. When we hover over the stacked bar chart in~\autoref{fig:vehicle}(a), we observe that safe and borderline cases account for 47.16\% and 34.54\% of the training set, respectively. This is significantly different in the distributions of data created with lower values for the number of neighbors (e.g., 5). In summary, the visual analysis guides us in picking all the aforementioned parameters. \linebreak

\textbf{Tuning the undersampling based on exploratory data analysis.} 
After selecting the projection (which results in a specific distribution of data types), we decide to apply the OSS undersampling algorithm. Nevertheless, the default settings cause the van class to disappear completely, thus the predictive performance gets extremely penalized (see step 1 in \autoref{fig:teaser}(h), right). We pick the highest available \emph{seeds} parameter to consider more points except for the minority class. The algorithm suggests 196 instances to be removed (step 2), as illustrated in~\autoref{fig:vehicle}(c). It seems from the projection that our previous setting does not capture several buses while being safe to remove examples. Therefore, we decrease the parameter to 125, half of the prior selection. The effect is that 329 are currently suggested for removal (step 3), as depicted in~\autoref{fig:vehicle}(d). This action accomplishes our initial goal, but 7 regional points are about to be undersampled. As we should be very careful when deleting rare cases, we further explore this group of points in the table heatmap view (\autoref{fig:vehicle}(e)). It is observable that the instance with ID 486 is separated from the others mainly due to the \emph{HollowsRat} feature, while instance 631 is different because of a low value in the \emph{MaxLensAspRat} feature. We decide not to exclude rare cases with such high variance because they may be part of our test set (step 4). The new suggestion after excluding the rare points is visible in~\autoref{fig:vehicle}(f). Another critical category of data types is the outliers. From all outliers in the last projection, four are proposed for deletion by the oversampling algorithm. Only 1 out of the 4 points appears marginally problematic with prominent confusion between car and van classes, as depicted in the inverse polar chart (see~\autoref{fig:vehicle}(g)). Since the majority of points is safely predicted correctly, we decide to keep the outliers in the training set. After this step, 256 out of the 634 points are getting removed (steps 5 and 6).

\textbf{Deciding to oversample all types except outliers.}
To understand if a new round of undersampling would be beneficial, we activate the OSS algorithm again with the same settings (step 7). However, the outcome is to decrease the relatively safe population that much, so that the result is becoming worse. Therefore, we disable the algorithm and stop the undersampling phase (step 8). Moving on to the oversampling phase, we aim at utilizing SMOTE to generate artificial points for increasing the number of instances in the underrepresented classes. The oversampling of all data types reduces both Balanced Accuracy and F1-score (step 9 in~\autoref{fig:teaser}(h)). From~\autoref{fig:vehicle}(f), we can understand that several problematic outliers are not considered for removal at all by the OSS algorithm during the previous phase. In particular, four outliers are predicted as vans while they belong to the car class according to the ground truth, as shown in~\autoref{fig:vehicle}(g), at the bottom-left region. The oversampling algorithm should not eternalize this confusion. Consequently, we choose to exclude all 6 outliers from the pool of instances in order to primarily generate safe and borderline instances for the van and bus classes (cf.~\autoref{fig:teaser}(a)). The resulting distribution of points achieves our goal (see~\autoref{fig:teaser}(c)) and leads to an improvement in the overall predictive power (step 10).

\textbf{Tracking the process and evaluating the results.} 
To verify our sampling execution actions, we continuously monitor the process through the Sankey diagram, as shown in \autoref{fig:teaser}(g). From this representation, we acknowledge that the population of safe instances decreased drastically when the undersampling was executed. The manual undersampling and oversampling processes (described previously) led to the best predictive result we managed to accomplish, with 9 confused test instances (7 of them belonging to the car class, as presented in \autoref{fig:teaser}(h), left). From the horizontal bar chart in \autoref{fig:teaser}(h), right, the performance difference in each step suggests that using directly the automated sampling algorithms led to worse results (cf. steps 1 and 9). With the help of \textsc{HardVis}, we managed to improve, even more, both Balanced Accuracy and F1-score by approximately $+2\%$. To sum up, our VA system guided us in systematically setting the parameters of the sampling algorithms and applying them in subsets of the data throughout the various rounds of undersampling and oversampling. As pointed out by the experts in~\autoref{sec:eval}, this would have been (almost) impossible without direct human intervention.

\section{Evaluation} \label{sec:eval} {%
  We performed online, semi-structured interviews with five independent experts to gain qualitative feedback on our system's usefulness, using the procedure described in prior works~\cite{Ma2020Explaining,Xu2019EnsembleLens}. The first ML expert (\textbf{E1}) is a full professor with a PhD in computer science. He has 15 years of experience with ML, and he is head of the natural language processing (NLP) group at his university. 
The second ML expert (\textbf{E2}) is a full professor in ML and data science addressing mainly challenges in humanities. He has worked with ML for the past 30 years, and he holds a PhD in applied mathematics. The third ML expert (\textbf{E3}) is an assistant professor working with ML and deep learning, with 7 years of experience in ML. His PhD is in media technology. The fourth ML expert (\textbf{E4}) is a postdoc also focusing on ML and deep learning, and she has 8 years of experience in ML. Finally, the fifth ML expert (\textbf{E5}) is a postdoc with 20 years of experience in ML. The latter two experts have PhDs in computer science.
\textbf{E1} was the only one who reported a colorblindness issue (deuteranomaly), but he affirmed having no problem perceiving correctly the specific color combinations we used in \textsc{HardVis}.
\hl{Each interview lasted about 1 hour and 15 minutes, and the interviews were structured as follows:
(1) introduction of the primary objectives of \textsc{HardVis}, including the analytical tasks and design goals of~\autoref{sec:goals}; (2) presentation of the functionality of every visualization and interaction with the system using the \emph{iris flower} data set (as in~\autoref{sec:system}); and (3) explanation of the steps taken to arrive at the results in~\autoref{sec:case}.}
We asked the participants to freely comment on anything. Their responses are summarized below.

\textbf{Workflow.}
\textbf{All experts} agreed that \textsc{HardVis'} workflow is well designed and reasonable from their perspective. They characterized the workflow as straightforward and aligned with respective fully-automated sampling processes. \textbf{E1} and \textbf{E2} repeatedly commented positively upon our systematic and fine-grained approach that they have never seen before in all those years of developing new and deploying already existent ML models. ``The offered granularity of undersampling and oversampling is exceptional, i.e., the fact that several phases can be applied in a row and for different subsets of the data space is something that I believe is almost impossible to accomplish without such a tool'', said \textbf{E1}. \textbf{E2} underlined the clear benefit of controlling the automatic algorithms' suggestions since blindly following them could overfit the training set (and hurt generalization). He then stated that letting users be completely free to remove or generate artificial instances manually could probably harm the predictive performance similarly. Thus, \textbf{E2} found that our tool combines the best of both worlds.

\textbf{Visualization and interaction.}
The promising findings we were able to obtain with the help of our VA system in the usage scenario of~\autoref{sec:case} amazed \textbf{E3} and \textbf{E4}. While using the same value for the number of neighbors parameter and the k-value for the distribution of data types, \textbf{E3} appreciated that the k-value could still be adapted freely, as illustrated in~\autoref{fig:iris-under}(c). The most intuitive visualization according to \textbf{E2}, \textbf{E4}, and \textbf{E5} was the box plots view (\autoref{fig:iris-under}(f)) which was found exceptionally well-linked with the UMAP projection (\autoref{fig:iris-under}(d)). Especially with this view, \textbf{these experts} 
were able to understand the decisions we made in~\autoref{sec:system} and~\autoref{sec:case}. The inverse polar chart (cf.~\autoref{fig:iris-under}(i)) was the most confusing view at first. However, after a careful explanation from our side, \textbf{all experts} understood its meaning and claimed this visualization was the most novel visual representation of our tool. Since the same encoding as with the UMAP projection makes this view intuitive, they were able to inspect the instances immediately with low predicted probability (and with which specific class) from the eyes of the model. An interesting suggestion by \textbf{E2} was to visualize the KNN-graph for a particular instance when users hover over a specific point/instance. Although \textsc{HardVis} already enables users to make justifiable actions by exploring all training instances from both global and local perspectives, this recommendation could be seen as an extra validation step for the projection. He also mentioned that for a more unsupervised-focused approach, the main color of the projected points could show the data types, and the outline of points could be used for the ground truth labels (if there are any). Despite that, \textbf{E3} and \textbf{E4} thought that with the current color scale, the focus is on unsafe cases, which could decrease the model's accuracy if they are removed before or without reasoning about them at all.

\textbf{Limitations identified by the experts.}
\textbf{E1} and \textbf{E2} were concerned about the \emph{scalability} of the system. The former concentrated on the problem of visualizing hundreds of features, while the latter on the exploration of more than three classes. \textbf{E1} acknowledged that the box plots and the table heatmap view are interactive with zooming and panning functionalities, which could partially address this issue. Also, the feature importance could be useful for deciding which features are not informative for a provided data set to exclude them beforehand. Regarding the second issue, the main bottlenecks are the inverse polar chart and the extensive use of colors. The proposed visualization could be further improved to scale with more than three classes by using advanced RadViz-based approaches~\cite{Piazentin2015Concentric,Thanh2014Reordering}. Furthermore, \textbf{E2} noted that multi-class classification problems could be resolved as being binary due to the one-vs-rest strategy. \textbf{E5} proposed to deploy \textsc{HardVis} in a cloud server supporting parallel processing to improve further the \emph{efficiency} of the system. \textbf{E1} and \textbf{E3} mentioned that heavily modifying our VA system is inevitable in case we would like to extend it to \emph{other types of data}, e.g., image or NLP data sets that consist of non-interpretable features such as pixels and word vectors. However, they completely agreed that this was not our original intention. \textbf{E1} stated that non-expert users or even domain experts could find it difficult to operate \textsc{HardVis} and be advised by all visualizations concurrently, despite the views being logically positioned in a single window. Therefore, as an improvement of \emph{generalizability to other target groups}, he proposed to separate the views in different tabs depending on the certain domain problem at hand and the users' prior experience to reduce the cognitive load. However, for ML experts, this deep level of granularity and the guidance received from the tool are necessary for making decisions. Finally, \textbf{E3} described that as with any other VA tool and ML model in general, the \emph{quality of the data set} would probably affect negatively the capability of the tool to explore a complex and low-quality data set to the point that it could be challenging to improve the predictive performance. A preprocessing phase that handles missing values and wrangles the data could alleviate this problem. We plan to work on methods to surpass such limitations. 

\textbf{Overall assessment.} The provided feedback was encouraging and in favor of \textsc{HardVis} compared to employing automatic approaches. \textbf{All experts} were confident about the benefits of using our VA system.

\section{Discussion} \label{sec:dis} {%
  In this section, we discuss the visual design and overall limitations of our approach as well as the current implementation.

\subsection{Visual Design} \label{sec:design}

Here, we elaborate further on the key design concepts of our VA system that were presented in~\autoref{sec:system}.

\textbf{Familiarity with the prevalent types of data visualization.}
The visual representations used are intentionally simple but form a powerful system when combined. Specifically, the benefits originate from the identification of areas where sampling strategies should be applied with guidance across the entire process. Similar to the user profile selected for the ML experts that participated in our interview sessions, we deem that the users of our tool would have worked with box plots, bar charts, tabular representations, and visual embeddings in the past. Therefore, there may be a gradual learning curve relevant to the familiarity with the visualizations. Two exceptions could be the Sankey diagram and the inverse polar chart. \hl{The former is for keeping track of their actions (usually studied under the term \emph{provenance} in visualization~\cite{Xu2020Survey}). A simpler alternative we considered is a log list of user's actions being registered in each step, as well as empowerements of this representation with highlighted text. However, it would capture too much space for a view that can be deemed as optional, especially since the Sankey diagram is not crucial during the exploration and analysis phases (i.e., before either undersampling or oversampling take place).} The latter representation needs to be learned but can be a game-changer for finding instances of confusion with a particular class and observing the distribution of SBRO types from the perspective of the ML model, as already mentioned in~\autoref{sec:eval}. \hl{As a straightforward alternative, we tried out a multi-class confusion matrix. However, it only provides aggregated information and fails to use the same visual encodings as the main view (see below).}

\textbf{Commonality in the visual encoding and color scales.}
Throughout the whole \textsc{HardVis} system, the visual encodings propagate from one view to the others. For example, the common grayscale denotes the four distinct types of instances in all views. Tightly connected views---such as the UMAP projection and the inverse polar chart---share identical encodings, i.e., label class mapped to filled-in color, data type as outline color, and US/OS represented with symbols. The inverse polar chart is compact and uses the available space effectively due to its inherent design; it spares more area for the misclassified instances. For the table heatmap view, the diverging color scale emphasizes the extreme values and allows users to notice more differences on the left- and right-hand sides of the middle point, with five colors having the same origin. For example, this middle point is crucial for the \emph{breast cancer} data set, because instances with values closer to 1 for all features should be classified as malignant, while samples with values around 0 should be benign cancer. Finally in this view, hovering over a specific cell interaction partly resolves the ambiguity problem introduced due to distributing the normalized values into 10 distinct bins.

\subsection{Limitations} \label{sec:addlimit}

In the following, we acknowledge limitations we have discovered for our system (beyond those mentioned in~\autoref{sec:eval}), which imply prospective future developments.

\textbf{Scalability for a large number of instances and features.}
In general, the number of instances and features that can be visually expressed with our approach has no intrinsic limit. Collaris and van Wijk~\cite{Collaris2022StrategyAtlas} found that usually the top 10--20 features were impactful for the tabular data sets they experimented with. For hundreds of features, it would be cognitively demanding for a human to analyze the influence of all these features at different levels of granularity. The methodology that might be used is first to limit the space under inspection using an additional preprocessing phase in the pipeline before employing \textsc{HardVis} for a deep analysis of features, as already stated by an ML expert in~\autoref{sec:eval}. Collaris and van Wijk~\cite{Collaris2022StrategyAtlas} also limited the number of instances to 5,000 in order to prevent overplotting issues in their projection-based view. Arguably, similar constraints should apply to our tool, especially for the UMAP projection and the inverse polar chart view. However in our case, zooming and panning functionalities implemented for both views can partly solve this problem along with overlap removal strategies that could be helpful~\cite{Hilasaca2019Overlap,Yuan2021Evaluation}.
Regarding the table heatmap view, it is mostly useful for comparing a group of instances after a lasso selection has been performed. Additionally, we have the box plots that offer an overview first and scale better to many more instances.

\begin{table}[tb]
\begin{threeparttable}
\caption{Time taken to complete each activity of the sampling process for all use cases. The completion time is expressed in \emph{minute:second} format. Please note that for the \emph{iris flower} data set, the undersampling time refers to two consecutive rounds.}
\label{time}
\setlength\tabcolsep{0pt} 
\begin{tabular*}{\columnwidth}{@{\extracolsep{\fill}} l ccc}
\toprule
     Data set &
     \multicolumn{3}{c}{Sampling process} \\
\cmidrule{2-4}
     & Data types & Undersampling & Oversampling \\
\midrule
     Iris flower & 0:45 & 2:57 & 1:06 \\
     Breast cancer & 1:53 & 6:52 & - \\
     Vehicle silhouettes & 3:29 & 8:58 & 5:12 \\
\bottomrule
\end{tabular*}
\smallskip
\scriptsize
\end{threeparttable}
\end{table} \vspace{-1mm}

\textbf{Other kinds of data sets.}
Despite the vast range of application domains covered with all our use cases, \textsc{HardVis} has merely been evaluated with structured tabular data consisting of numerical values~\cite{Shwartzziv2021Tabular}. We want to enable other data types in the future. Nevertheless, the features of each data set under investigation should be meaningful, because we focus on human expertise and knowledge to resolve problematic situations where essential instances for the generalizability of unseen data are being considered for deletion and to avoid the generation of artificial samples that negatively impact the predictive performance of the model. \hl{Overall, since} our prototype tool is a proof-of-concept, the system's workflow and theoretical contributions are generalizable in this respect.

\textbf{Target group.}
The primarily targeted users that would gain the most from adopting our approach 
are ML experts. We suppose that they understand the fundamentals of their data sets and know how to interpret common visual representations, but they require additional assistance with the sampling procedure. As evident from~\autoref{sec:eval}, the five ML experts who participated in our 1-hour and 15-minute interview sessions were able to grasp the main concepts and operate \textsc{HardVis}. Another potential here is to create a more basic version of our tool, geared explicitly for ML developers and even inexperienced ML users with a low level of visualization literacy.

\textbf{Completion time for each activity.} 
The frontend of \textsc{HardVis} has been developed in JavaScript and uses Vue.js~\cite{vuejs}, D3.js~\cite{D3}, and Plotly.js~\cite{plotly}, while the backend has been written in Python and uses Flask~\cite{Flask} and Scikit-learn~\cite{Pedregosa2011Scikit}. \hl{More technical details are made available on GitHub~\cite{HardVisCode}.} All experiments were performed on a MacBook Pro 2019 with a 2.6 GHz (6-Core) Intel Core i7 CPU, an AMD Radeon Pro 5300M 4 GB GPU, 16 GB of DDR4 RAM at 2667 Mhz, and running macOS Monterey. By taking into account the specifications of the computer, we recorded the total wall-clock time dedicated to completing the sampling process for each data set (see~\autoref{time}, rows). For the time reported, we aggregate both the computational analysis and the execution of the user's actions, as described in Sections~\ref{sec:application} and~\ref{sec:case}. \autoref{time} columns map the time for each activity of the sampling process (i.e., distribution of data types, undersampling phase, and oversampling phase). In particular, as the number of instances and features to be examined grows, so does the time necessary to compare alternative options and finalize the user-defined actions. Unsurprisingly, the undersampling phase took the longest in all situations, followed by the oversampling phase, and lastly the distribution of data types. Depending on the quantity and importance of the extracted patterns, these values might become rather different. In general, the rendering time after a major user's action is restricted to a couple of seconds for all the data sets we tried. To sum up, the efficiency of \textsc{HardVis} could be increased in various ways, as explained before.

\section{Conclusion} \label{sec:con} {%
  In this paper, we developed \textsc{HardVis}, a VA system that uses hardly-configurable undersampling and oversampling techniques to handle instance hardness. As part of an intensively iterative process, multiple coordinated views assist users in defining an ideal distribution of data types, undersampling particular safe for removal samples, and oversampling others. Additionally, it facilitates the exploration of algorithmic suggestions using a variety of visual clues to confirm non-harmful removal or addition proposals. Finally, our VA approach is ideal for dealing with the instance hardness and class imbalance challenges because it makes the entire process adjustable and more transparent. 
The effectiveness of \textsc{HardVis} was investigated using real-world data sets, which revealed an increase of trustworthiness and in performance due to removed and synthetically-generated instances.
The workflow and visualizations of our system received positive feedback from experts suggesting that such in-depth sampling would be impossible without our tool. They also assisted us in identifying the existing limitations of \textsc{HardVis}, which we are considering as future research directions.
  
\section*{Acknowledgements}
This work was partially supported through the ELLIIT environment for strategic research in Sweden.

\bibliographystyle{eg-alpha-doi}

\bibliography{HardVis}

\end{document}